\documentclass[10pt,twocolumn,letterpaper]{article}

\usepackage{iccv}              

\usepackage{algorithm}
\usepackage{algorithmic}
\usepackage{graphicx}
\usepackage{subcaption}
\usepackage{multirow} 
\usepackage{natbib}
\usepackage{makecell}
\definecolor{revised}{RGB}{0,0,0}
\definecolor{iccvblue}{rgb}{0.21,0.49,0.74}
\usepackage[pagebackref,breaklinks,colorlinks,allcolors=iccvblue]
{hyperref}

\def\paperID{7595} 
\def\confName{ICCV}
\def\confYear{2025}

\title{ Physically Real-time Infrared Attack against Optical Flow Estimation Networks }
\author{Shen You\\
City University of Hong Kong\\
Tat Chee Avenue, Kowloon, Hong Kong SAR\\
{\tt\small shenyou2-c@my.cityu.edu.hk}
\and
Wei Jiang\\
University of Electronic Science and Technology\\
Chengdu, Sichuan, China\\
{\tt\small weijiang@uestc.edu.cn}
\and
Jiarui Liu\\
University of Electronic Science and Technology \\
Chengdu,Sichuan,China\\
{\tt\small jiarui.liu@std.uestc.edu.cn}
\and
Yijian Ye\\
University of Electronic Science and Technology\\
Chengdu,Sichuan,China\\
{\tt\small 202321090116@std.uestc.edu.cn}
\and 
Qiuzhen Lin\\
Shenzhen University\\
Shenzhen,Guangdong,China\\
{\tt\small qiuzhlin@szu.edu.cn}
\and 
Xiangtao Li\\
Jilin University\\
Changchun,Jilin,China\\
{\tt\small lixt314@jlu.edu.cn}
\and 
Ka-Chun Wong*\\
City University of Hong Kong\\
Tat Chee Avenue, Kowloon, Hong Kong SAR\\
{\tt\small kc.w@cityu.edu.hk}
}

\begin{document}


\maketitle
\begin{abstract}
With the promising performance of deep neural networks on image-based tasks, different real-world applications such as autonomous driving and motion detection have become increasingly mature and relevant to human lives. In particular, Optical Flow Estimation Networks (OFENs), as upstream models, play a critical role in different domains. Its outputs are heavily assumed and adopted for different downstream tasks, and it is essential to test its robustness to prevent safety accidents.

We present an approach for real-time attacks on OFENs in the physical world, leveraging infrared lights for their stealthiness.
By generating a large number of Adversarial Examples (AEs) in advance, our approach computes AEs in real time and dynamically displays them, which allows our method to facilitate precise and targeted attacks without modifying the victim system.
Unlike previous digital-to-physical attack techniques, our method directly attacks victim models within the physical world, thereby overcoming the limitations associated with the ineffectiveness of AEs. 
Experimental results demonstrate the efficacy of our approach in compromising OFENs across diverse lighting conditions, varying object motion velocities, and different object placements, ultimately impairing the network’s ability to accurately estimate optical flow. 
\end{abstract}
\section{Introduction}
\textcolor{revised}{
Optical Flow Estimation Networks (OFENs) are proposed to estimate the motion between two image frames. It has the capability to identify and track moving objects under a static background by analyzing the changes in  optical flow. These capabilities have enabled OFENs to be widely adopted in various applications, such as motion recognition \cite{chamorro2006new}, behavior analysis \cite{sivaraman2013looking}, vehicle tracking \cite{haag1999combination} and obstacle avoidance \cite{song2001fast}. 
For example, Capito et al. \cite{capito2020optical} propose an artificial potential field, known as a visual potential field, from a sequence of images using sparse optical flow. This field is then utilized in conjunction with a gradient tracking sliding mode controller to navigate the vehicle to its destination while avoiding obstacles.
Typically, OFENs are considered an upstream task model where its outputs serve as the inputs for other downstream models. Consequently, it significantly influences the performance of downstream models. Therefore, the robustness of OFENs in autonomous driving has garnered considerable attention.
}

\textcolor{revised}{
However, as a type of Deep Neural Networks (DNNs), OFENs inherit similar vulnerabilities to adversarial perturbations \cite{narodytska2017simple,croce2019sparse}, which can significantly disrupt model outputs. This poses a unique challenge in autonomous driving as erroneous predictions can lead to severe consequences for the safety of drivers and passengers, where OFENs' outputs are critical for downstream decision-making systems.
\begin{figure}[t]
\centering
 \begin{subfigure}[b]{1\textwidth}
\includegraphics[width=0.256\textwidth]{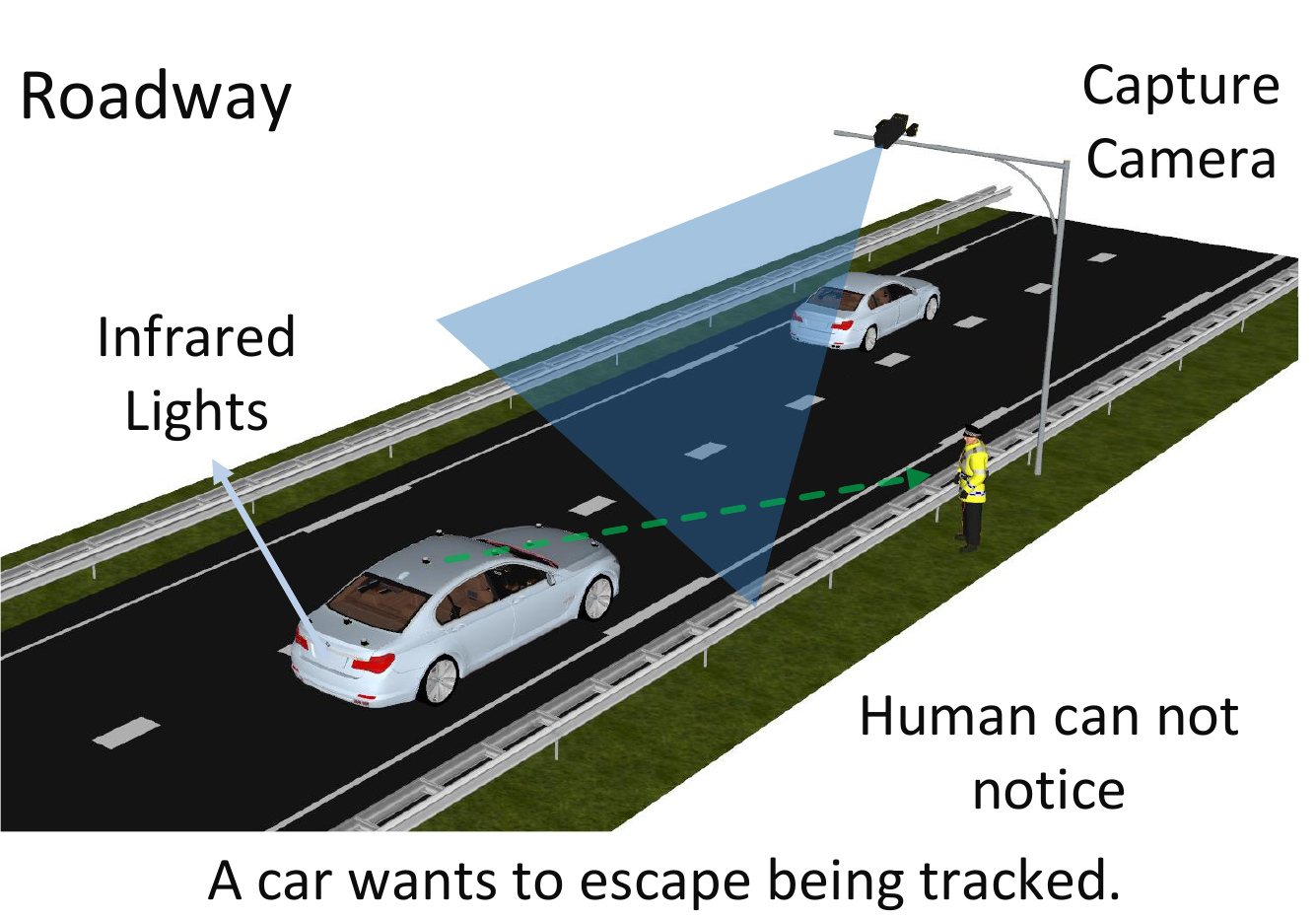}
\includegraphics[width=0.248\textwidth]{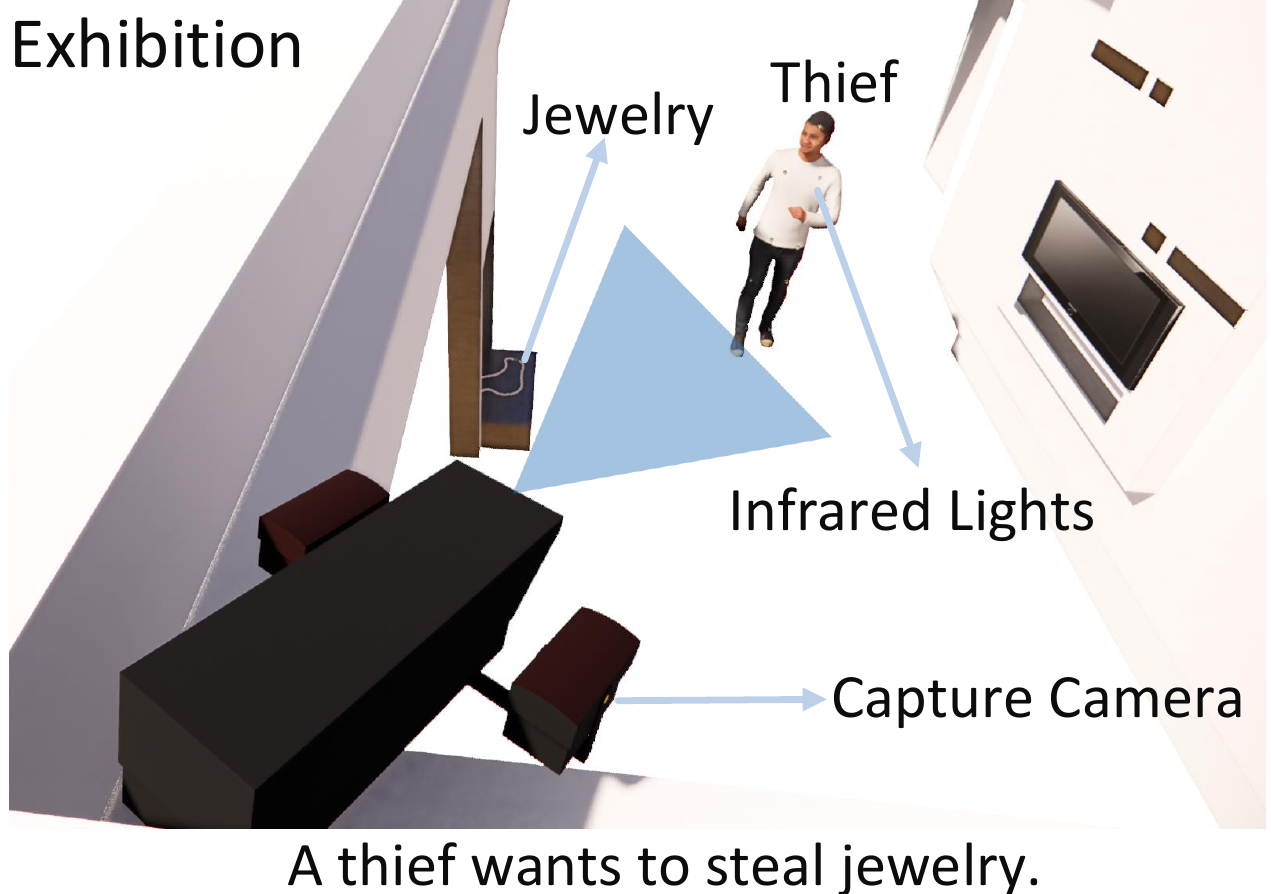}
\end{subfigure}
\caption{The application infrared attack.} 
\label{fig:pre_attack}
\end{figure}
Despite the extensive research on Adversarial Examples (AEs) in the physical world \cite{wang2021can,eykholt2018robust,zheng2023robust,bhupathiraju2024vulnerability}, there are several critical limitations. 
First, most studies evaluate the performance in the digital domain, which limits their practicality due to the complex transformations required for transferring these attacks to the physical world. 
Second, some physical attacks \cite{liu2019perceptual,wang2023rfla}  do not account for the differences between the digital and physical domains; they simply print AEs generated in the digital world. However, cameras inevitably capture noise, which can degrade the effectiveness of these AEs. 
Third, they do not consider the movement of target object \cite{brown2017adversarial,guo2024invisible}, which leads to failures when the target object changes its angle or position, particularly in the context of attacking OFENs. 
Finally, most attacks rely on visible patches or lights \cite{hu2021naturalistic,lee2019physical,duan2021adversarial}, which can easily be noticed by humans, thus violating the principle of imperceptibility in AEs. 
}

\textcolor{revised}{
To address the aforementioned limitations, we propose a physically real-time infrared light attack method. 
Figure \ref{fig:pre_attack} shows the application of our method. The left figure shows the attacker paste the infrared lights on the surface of car, allowing it been monitored and escape from traffic police. The right figure shows a thief wants steal jewelry and leverage infrared lights to spoof OFEN. Since most cameras are designed to work in low-light environments, they are usually not equipped with infrared filters, which gives good physical conditions for our attacks.
}
Our contributions can be summarized as follows:
\begin{itemize} 
\item A real-time adversarial attack method is proposed to shift the attack paradigm from digital-attack and physical-deploy to physical-train and physical-deploy. This approach significantly reduces the likelihood of AE failures.

\item Infrared lights are proposed to reduce the probability of AEs being detected because of the invisibility to the human eyes, enabling our attacks to be stealthy and imperceptible.


\item Three losses are introduced to effectively guide the generation of AEs for attacking OFENs. Experimental results show that our method achieves robust attack performance across multiple scenarios.
 
\end{itemize}

\section{Related Works}
In the following, we mainly focus on the development of OFENs and related attack work in this field.

\subsection{Optical Flow Estimation}
The OFENs are used to track the motion of the target object, which is based on the following three assumptions:
\begin{itemize}
\item Brightness Constancy: In adjacent frames, the brightness of the same object remains unchanged.
\item Time Continuity Assumption: The interval between adjacent video frames is very small, or the motion of objects between adjacent frames is relatively small. This implies that the motion of the object is continuous between adjacent frames without sudden jumps.
\item Spatial Coherence: Pixels in the same sub-image have the same motion. This means that the motion of pixels in the same sub-image between adjacent frames is coordinated and consistent.
\end{itemize}

The development of deep learning on optical flow estimation networks can be categorized into three stages as follows:
In 2015, Dosovitskiy et al. \cite{dosovitskiy2015flownet} introduced deep learning methods to estimate optical flow, constructing convolutional neural networks capable of solving the optical flow estimation problem as a supervised learning task. They proposed two main network architectures: FlowNetSimple, which features a straightforward encoder-decoder structure, and FlowNetCorr, which first extracts features for each frame and then uses a correlation layer to compute the similarity between the features of the two frames. 

In 2018, Sun et al. \cite{sun2018pwc} introduced PWC-Net,  utilizes three key components: an image pyramid, warping, and a matching cost volume (similar to the correlation computation in FlowNet). The warping and matching cost volume computations do not require training parameters, thereby reducing the overall number of model parameters. PWC-Net achieves real-time optical flow estimation on 1024×436 resolution videos, processing up to 35 frames per second with only 0.06 times the number of parameters compared to FlowNet2.0. Subsequently, Teed et al. \cite{teed2020raft} introduced Recurrent All-Pairs Field Transforms (RAFT) in 2020, a novel deep network architecture for optical flow estimation. RAFT extracts per-pixel features, constructs multi-scale 4D correlation volumes for all pairs of pixels, and iteratively updates the flow field through a recurrent unit that performs lookups on these correlation volumes.

\begin{figure*}[tb]
 \centering
 \begin{subfigure}[b]{0.48\textwidth}
 \includegraphics[width=\textwidth]{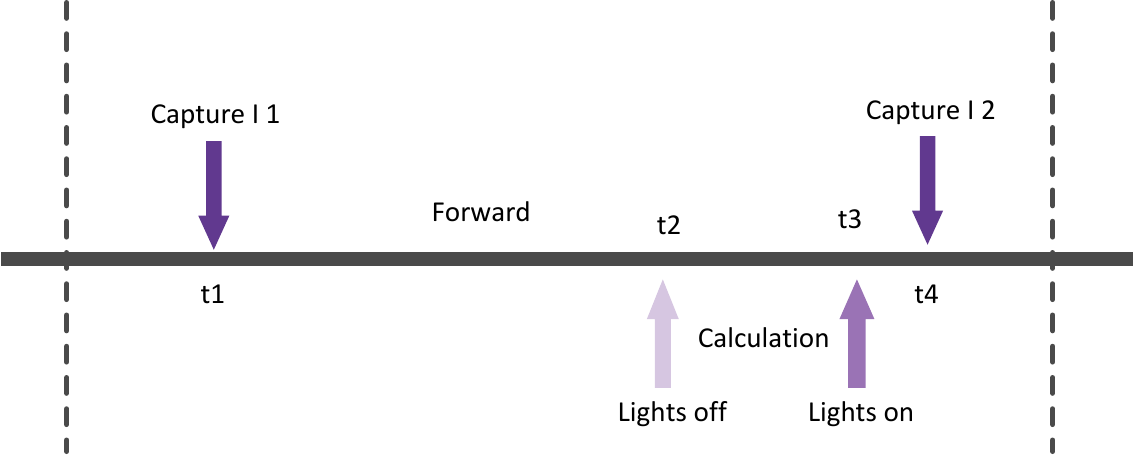}
 \caption{Timeline}
 \label{fig:timeline}
 \end{subfigure}
 \hfill
 \begin{subfigure}[b]{0.50\textwidth}
 \includegraphics[width=\textwidth]{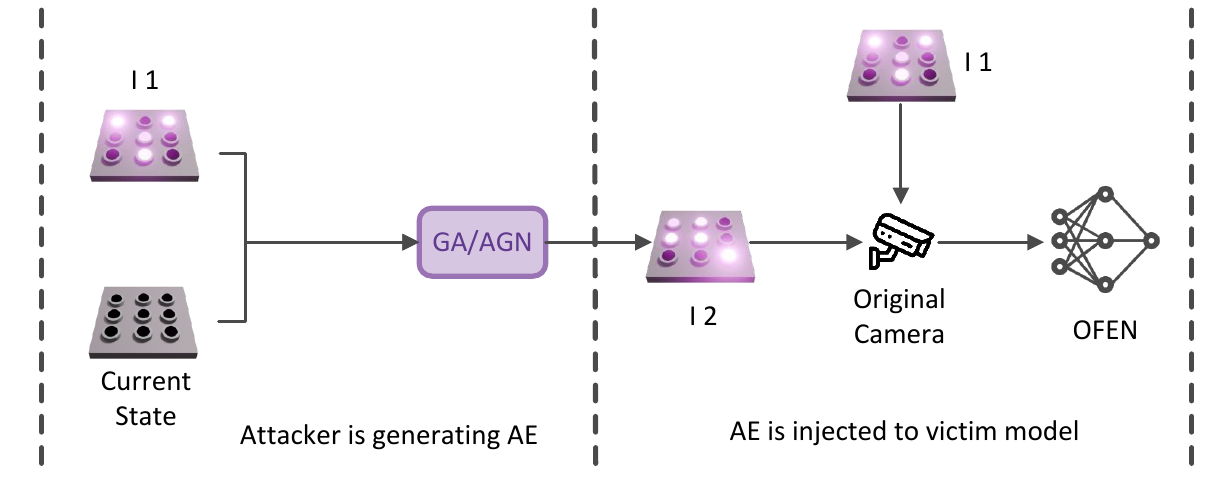}
 \caption{Sequence Chart}
 \label{fig:flowchart}
 \end{subfigure}
 \caption{(a) Timeline and (b) Sequence Chart of the OFEN attacking process with AEs in the physical world.}
 \label{fig:over_all}
\end{figure*}

\subsection{Physical Attacks for OFENs}
In recent years, there are some research on attacking OFENs. Ranjan et al. \cite{ranjan2019attacking} extended adversarial patch attacks to optical flow networks and found that corrupting a small patch, covering less than 1\% of the image size, can significantly affect optical flow estimates.
Abhiram et al. \cite{gnanasambandam2021optical} propose an adversarial attack in the physical world aimed at fooling image classifiers without physically touching the objects. Their method uses a low-cost projector to alter the appearance of the target objects.
Guo et al. \cite{guo2024invisible} present an attack that uses light-emitting diodes and exploits the camera's rolling shutter effect to create adversarial stripes in the captured images to mislead traffic sign recognition. 

Although the existing research achieved remarkable success, it still has some fatal problems. 
The methods used in the digital domain are often difficult to directly migrate and apply to the physical world and  different environmental conditions such as lighting, distance, and sensor viewing angle, the image captured by the sensor and input into the model will differ from the original image. 
Additionally, fabrication errors pose a significant challenge. To fabricate perturbations based on theoretical calculations, every pixel must be accurately printed using existing equipment, such as adversarial patches. Due to the limitations of the color gamut of modern printing equipment, some theoretically calculated colors may not be fully or accurately reproduced. 
\section{Problem Formulation} 

Optical flow $\mathbf{O} \in \mathbb{R}^{H\times W \times 2}$ indicates the motion of target object where $\mathbf{O}(x,y)=[\Delta x, \Delta y]^T$. For a real-time frame flow $\mathbf{I}^{i} \in \mathbb{R}^{H\times W \times C}$ with three channels (rgb) at time $t=i$, OFEN $\mathcal{F}(\cdot,\cdot)$ takes the previous frame $\mathbf{I}^{i-1}$ and current frame $\mathbf{I}^i$ to estimate optical flow $\mathbf{O} \in \mathbb{R}^{H\times W \times 2}$. The relationship between optical flow $\mathbf{O}$ and frame $\mathbf{I}$ is shown as Eq. \ref{eq:optical_flow} 
\begin{equation}
\frac{\partial \mathbf{I}}{\partial x}\Delta x + \frac{\partial \mathbf{I}}{\partial y}\Delta y + \frac{\partial \mathbf{I}}{\partial t} = 0
\label{eq:optical_flow}
\end{equation}
This differential form derives from the first-order taylor expansion under brightness constancy assumption. $x $ and $y $ represent the specific pixel coordinates in the frame, while $H $ and $W $ denote the height and width of the input frame, respectively. The optical flow $\mathbf{O}$ has two dimensions $(\Delta x,\Delta y)$ in each position, corresponding to the horizontal and vertical components of the flow.

Next, we need to define the perturbation for the infrared attack. Consider a scenario where $n $ infrared lights are attached to the surface of the target object, and each light has a normalized intensity range from 0 to 1. The perturbation $\mathbf{P} \in \mathbb{R}^n $ can be encoded as a vector with $n $ elements, where each element in $\mathbf{P} $ represents the intensity of the corresponding infrared light. Next, we define a non-derivative conversion function $\mathcal{C}(\cdot) $, which transforms the perturbation into the physical world. Specifically, Eq. \ref{eq:perturbation} shows the attacked optical flow at time $t = i $.
\begin{equation}
\mathbf{O}^{i}_* = \mathcal{F}(\mathcal{C}(\mathbf{P}^{i-1}, \mathbf{I}^{i-1}_*), \mathcal{C}(\mathbf{P}^i, \mathbf{I}^i_*))
\label{eq:perturbation}
\end{equation}
Then, we formulate the attack problem as follows:
\begin{equation}
\begin{split}
 \quad & \max \quad |\mathbf{O}^i_*-\mathbf{O}^i| \\
\text{s.t.} \quad & \mathbf{P}^i \in [0, 1], \quad t_i - t_{i-1} \leq \epsilon
\end{split}
\label{eq:problem}
\end{equation}
Here, $\epsilon$ represents the real-time constraint and the objective is to make the perturbed optical flow as different as possible from the original optical flow.

\section{Methodology}
In this section, we introduce a Real-time Infrared Light Attack (RILA). Our goal is to deploy a real-time attack to make sure the target object escapes being tracked by optical flow estimation network. Considering that the tracking camera captures frames every 30ms or less, making it necessary for the infrared light to react in a shorter period of time, we divide the whole attack process into two stages. Within the first phase, we build a training environment, which is the same as the real environment, and we use the GA algorithm to generate a large number of adversarial examples as training data. Within the second stage, we use the training data to train a perturbation generation network, which can generate adversarial examples in time.

\begin{itemize}
\item Training: An evolutionary algorithm is adopted to generate a large number of AEs as training data.
\item Attacking: Adversarial Generative Network (AGN) is used to generate AEs in real-time.
\end{itemize}

\subsection{Overall Framework}
\textcolor{revised}{
Fig. \ref{fig:over_all} (a) illustrates the timeline and sequence chart of the attack process. The arrows above the timeline represent the capture times of the camera, while the arrows below the timeline represent the actions of our attack. 
\begin{itemize}
 \item At $ t_1 $, the camera captures frame $ \mathbf{I}_1 $. 
 \item At $ t_2 $, we turn off the infrared lights, calculate the corresponding adversarial example and send it to the hardware controller.
 \item At $ t_3 $, the hardware controller displays the AE. 
 \item At $ t_4 $, the camera captures the frame $ \mathbf{I}_2 $ and uses frames $ \mathbf{I}_1 $ and $ \mathbf{I}_2 $ to estimate the optical flow.
\end{itemize}
}
\textcolor{revised}{
In the above analysis, there are some relationships between those times. Firstly, $\frac{1}{(t_4-t_1)}=\mathbf{R}$ means the frame rate of the camera, which is normally equal to 30 or 60. Besides, in the train stage, $ t_3 - t_2 $ can be longer because we can fix the position of the target object to provide sufficient time for the search process. In Attacking stage, $t_3-t_2$ must be very short to ensure real-time performance. 
Fig. \ref{fig:over_all} (b) shows the sequence chart of our attack process and how the AE disrupts the OFEN. We use GA (in Training stage) or AGN (in Attacking stage) to generate AE and use the hardware to display it. 
Once the AEs are presented, the attack at the current position is complete. The perturbed frame is then captured by the camera and fed into the target model, leading to the generation of an erroneous optical flow.
}
\subsection{GA for Non-Real Time Training} 
Owing to the convert function $\mathcal{C}(\cdot)$ is a physical process and is non-derivative, we cannot use the gradient-based optimization algorithm, so we use a genetic algorithm, which does not need gradients to generate AEs. The goal of GA is to generate an optimal light strength array $\mathbf{P}$ that constructs a perturbed frame according to Eq. \ref{eq:perturbation} to satisfy the objective defined in Eq. \ref{eq:problem}. 

\begin{figure}[htb]
\centering
\includegraphics[width=0.4\textwidth]{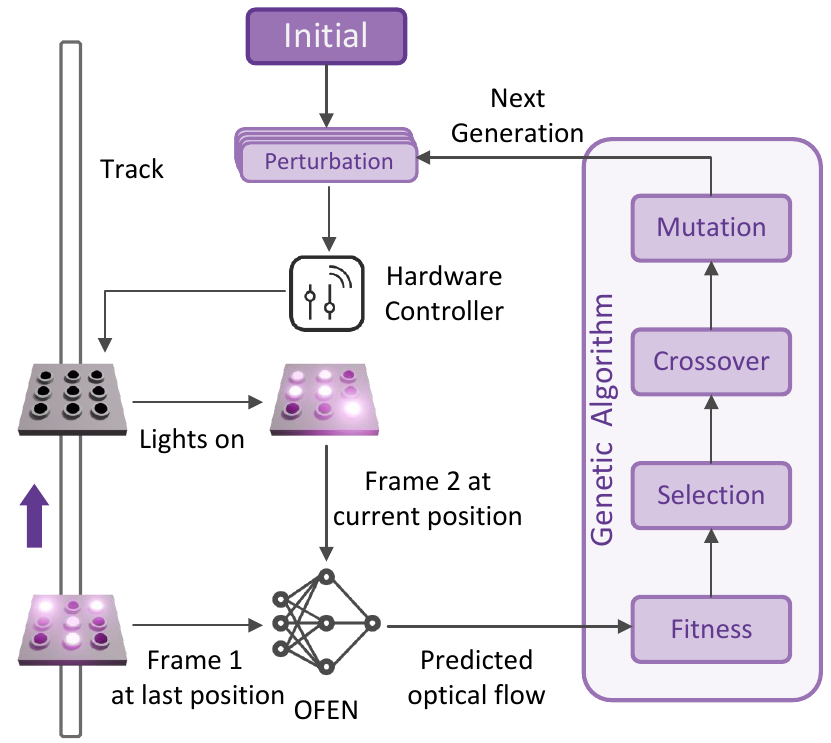}
\caption{The flowchart of AE generation with GA} 
\label{fig:GA}
\end{figure}
Fig. \ref{fig:GA} shows how we use GA to generate AEs. Firstly, we initialize a perturbation set, and then for each perturbation, we send it to the hardware controller. The hardware controller will set each infrared light at specific power to get the dirty frame, then the previous frame and the current dirty frame will be sent to OFEN to get the fitness. After we get the fitness of each perturbation, we need to construct the next generation of perturbations. This involves selection, crossover, and mutation. By iteratively applying these steps, the GA evolves the perturbations, gradually improving their ability to disrupt the OFEN. This process continues until the desired level of attack performance is achieved.

\subsubsection{Initial Perturbation Strategies}
Unlike other methods that use uniform distributions to initialize perturbations, we employ two preset initial plans based on the specific situation according to the feature of OFEN. We provide an example to explain how it works, considering the frames at $ t_1 $ and $ t_2 $. 
Since our attack primarily aims to break the brightness constancy assumption that the brightness of the same object remains unchanged in adjacent frames, we need to make the brightness between adjacent frames more different to break this assumption. Specifically, we need to increase the brightness in the current frame and then decrease it in the next frame. To control this process, we introduce a parameter $ \gamma $. If we are attacking odd frames, $ \gamma $ is set to 0.2; otherwise, $ \gamma $ is set to 0.8. The initial perturbation $ \mathbf{P}_i $ for each infrared light is given by:

\begin{equation}
\mathbf{P}_i = 0.1 \cdot\mathcal{U}() + \gamma
\label{eq:initial}
\end{equation}
Here, $\mathcal{U}()$ will return a uniformly distributed random value between 0 and 1. 

\subsubsection{Fitness Design}
The fitness function is crucial for guiding the evolution process and significantly impacts the performance of the evolutionary algorithm.
Although Eq. \ref{eq:problem} requires maximizing the difference between the optical flow and perturbed optical flow,  we choose the approximate approach for convenience, which is donated by these three sub-fitness terms. 

\begin{itemize}
\item Invisibility ($\mathfrak{L}_{Vis}$): This term is defined as $\mathfrak{L}_{Vis} = \frac{|\mathbf{O}^i|}{H\times W \times C} \in [0,1]$, measures the invisibility of the perturbation. When $\mathfrak{L}_{Vis}$ tends to one, it indicates that the optical flow is blank, meaning no moving object will be detected.

\item Imperceptibility ($\mathfrak{L}_{Per}$): This term, defined as $\mathfrak{L}_{Per} \in [0,1]$, it equals one minus the average variance of each sliding window, like in the convolution operation $1-\bar{{\mathcal{V}(i)}}$, $i$ means the sliding window. It measures the imperceptibility of the perturbation. A higher $\mathfrak{L}_{Per}$ value indicates that the optical flow is more chaos and unable to accurately perceive the motion, making it difficult for the model to identify the true movement.

\item Flash Loss ($\mathfrak{L}_{Flash}$): Based on previous experience, we introduce a flash loss term, $\mathfrak{L}_{Flash} = \frac{|\mathbf{P}^i - \mathbf{P}^{i-1}|}{n} \in [0,1]$. A large $\mathfrak{L}_{Flash}$ value indicates that brightness in adjacent frames is more different, which can decrease the victim model's performance.
 
\end{itemize}
Given these three sub-fitness terms, the overall fitness function is a typical multi-objective optimization problem and is defined as:

\begin{equation}
Fitness(P^i) = \alpha \times \mathfrak{L}_{Vis} + \beta \times \mathfrak{L}_{Per} + \mathfrak{L}_{Flash}
\label{eq:fitness}
\end{equation}
where $\alpha$, $\beta$, are weighting factors.


\subsubsection{Perturbations Update}

The update of perturbations in our evolutionary algorithm consists of three main genetic operators: selection, crossover, and mutation. We use those operators to generate the next generation.

\emph{\textbf{Selection Operator}}

A selection operator is used to retain domain genes into the next population. We adopt a tournament selection strategy similar to that used in NSGA-II. The process is divided into the following steps: 
Firstly, we randomly select a fixed number of perturbations from the current population. Then, among the selected perturbations, choose the one with the highest fitness score. Next, the individual with the best performance is designated as the 'winner' and may become part of the new generation. Finally, the above steps are repeated multiple times until the required number of perturbations is obtained.

\emph{\textbf{Crossover Operator}}

In the crossover operator, we combine two parent perturbations to create offspring. We select multiple fixed crossover points and exchange the perturbation values between the parents. The resulting offspring inherit a combination of traits from both parents, which can lead to new and potentially better solutions.

\emph{\textbf{Mutation Operator}}

In the mutation operator, we introduce small random changes to the perturbation values of the selected individuals. This helps to maintain diversity in the population and prevents premature convergence. Specifically, for each element in the perturbation vector $\mathbf{P}$, we apply a small random perturbation drawn from a uniform distribution. This ensures that the perturbation remains within the valid range [0, 1]. $\mathbf{P}_i' = \mathcal{U}()$, where $ \mathbf{P}_i' $ is the mutated perturbation value. 

\subsection{AGN for Real Time Attacking}
\textcolor{revised}{
Since the time-costing nature of GA is not suitable for real-time attacks, we design a generative network to generate AEs using the last perturbation to generate the next perturbation. In order to fulfill the real-time requirement, the AGN network must generate perturbations in a short period of time, so we use a simple network with a fully connected layer for perturbation generation. This model is also easy to deploy at an embedded system and the input of it is $\mathbf{P}^{i-1}$ and output is $\mathbf{P}^{i}$.
}

\textcolor{revised}{
\subsection{Pseudocode of Real-time Physical Attack}
\begin{algorithm}[htb] 
 \caption{Pseudocode of RILA} 
 {\large \textbf{Input:}} \emph{The attack parameters.} \\
 {\large \textbf{Output:}} \emph{The dirty optical flow.} 
 \\
 \textbf{{\emph{\textcolor[RGB]{0,0,0}{Generate Training Dateset}}}}
 \begin{algorithmic}[1]
 \WHILE { Target object is on the track} 
 \STATE Capture $\mathbf{I^{i-1}}$ from camera for time i-1.
 \STATE Set move instruction to hardware controller.
 \STATE $\mathbf{P^i}$ = $GA(\mathbf{I}^{i-1})$.
 \STATE Add $(\mathbf{P^{i-1}},\mathbf{P^{i}})$ as train data.
 \STATE Send $\mathbf{P}^i$ to hardware controller and display it
 \ENDWHILE
 \STATE Using data to train $ AGN$
\end{algorithmic}
 \textbf{{\emph{\textcolor[RGB]{0,0,0}{Physical Real-Time Attack }}}}
 \begin{algorithmic}[1]
 \WHILE { Target object is under tracking} 
 \STATE $\mathbf{P}^i$ = $ AGN(\mathbf{P}^{i-1})$.
 \STATE Send $\mathbf{P}^i$ to hardware controller and display it.
 \STATE Wait for $\mathbf{R}$ millisecond.
 \ENDWHILE
 \end{algorithmic}
 \label{alg:realtime_attack} 
\end{algorithm} 
Alg. \ref{alg:realtime_attack} shows the pseudocode for the real-time physical attack. In training stage, we mainly focus on the data collection. While the target object is still on the track, we capture the initial frame $ \mathbf{I}_1 $ at time $i-1$. Then, We send a move instruction to the hardware controller. By using the GA, we generate the optimal perturbation $\mathbf{P}^i$, and we add the pair $(\mathbf{P^{i-1}},\mathbf{P^{i}})$ to the training dataset. Next, we send the perturbation $\mathbf{P}^i$ to the hardware controller and display it. Finally, after collecting sufficient training data, we use this data to train the AGN.
In real-time attacking stage, we deploy a real-time attack. If the target object is still under monitoring,  we generate the perturbation $\mathbf{P}^i$ from $P^{i-1}$ at time $i$,  and we send the perturbation $\mathbf{P}^i$ to the hardware controller and display it. Next, we wait for $\mathbb{R}$ milliseconds to wait for the next iteration.
}

\section{Experiments}
In this section, experiments are conducted to evaluate the effectiveness of the proposed method. All experiments have been performed on an Intel(R) Xeon(R) Gold 5218 CPU at 2.30GHz, one GPU of NVIDIA GeForce RTX 2080Ti, nine 840nm infrared lights, and an ESP32 controller.

\subsection{Experiment Setup}
\subsubsection{Physical Test Environment}
To test RILA, we constructed a physical test environment consisting of five main components: a track, ambient lighting, attached infrared lights, camera, and a controller.
\begin{itemize}
 \item Track: The track is used to control the movement of the target object, allowing us to easily stop the object and wait for the GA to compute the AEs. 
 \item Ambient lighting: This component helps us evaluate the performance of RILA under different lighting conditions.
 \item Attached infrared lights: Infrared lights are attached to the surface of the target object and dynamically display the AEs.
 \item Camera: The camera captures 30 frames per second, which is used for estimating the optical flow.
 \item Controller: The controller manages the track, ambient lighting, and attached infrared lights.
\end{itemize}

\begin{figure}[t]
\centering
\includegraphics[width=0.37\textwidth]{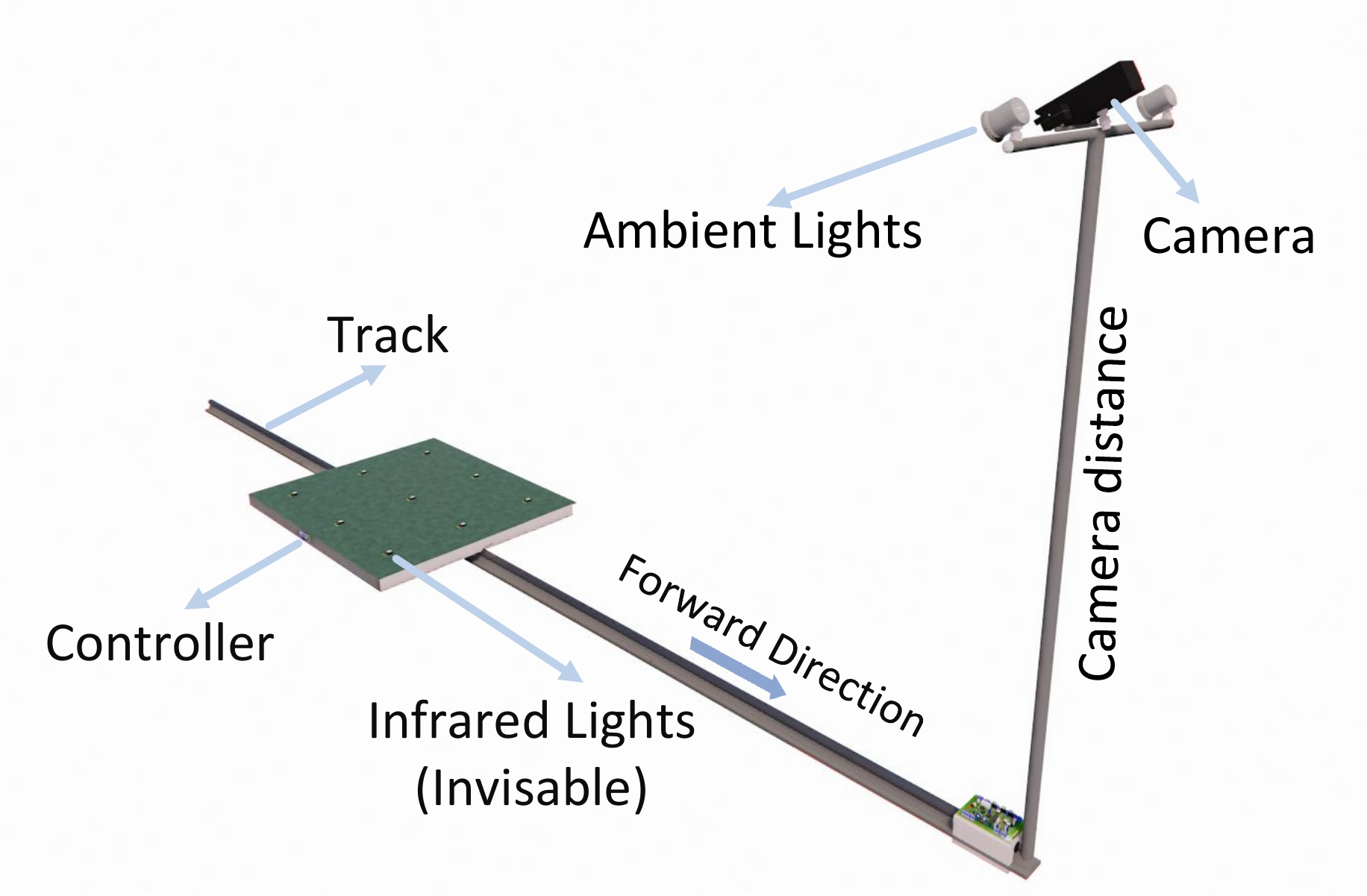}
\caption{The experiment environments.} 
\label{fig:exp_setting}
\end{figure}
The physical test environment is shown in Figure \ref{fig:exp_setting}. Considering that OFEN only detects the movement of pixels, not the change of shape and color of objects. So we embedded an infrared light on a plastic plate instead of a small car or a human body.

\subsubsection{Victim Model}
As for victim model, we choose the existing SOTA model RAFT \cite{teed2020raft} and another milestone model PWC-Net \cite{sun2018pwc}. 
As for comparison methods, because there are not any methods with the same physical test environments and we do not have the digital attack stage, we only test our methods in this work.
\subsubsection{Performance Metrics}
Two performance metrics are computed to evaluate the performance of RILA: Average Invisibility (AIV) and Average Imperceptibility (AIP). Higher AIV means the lower probability to be detected by downstream model. Higher AIP means it is hard to get the true motion of the output optical flow.

\subsection{Ablation Study}
In this section, we mainly focus on the validation of loss item, We test a version with all loss item against versions without individual loss item.
\begin{table}[htbp]
 \centering
 \caption{ Ablation Study Results.}
 \begin{tabular}{c@{\hspace{1pt}}c@{\hspace{7pt}}c@{\hspace{7pt}}c@{\hspace{7pt}}c@{\hspace{0pt}}}
 \toprule
 \multirow{2}[4]{*}{\makecell{Dis\\(cm)}} & \multicolumn{2}{c}{RAFT \cite{teed2020raft}} & \multicolumn{2}{c}{PWC-Net \cite{sun2018pwc}} \\
\cmidrule{2-5} & AIV & AIP & AIV & AIP \\
 \midrule
 N-Vis & .887 ± .007 & .974 ± .004 & .901 ± .011 & .980 ± .004 \\
 N-Per & .888 ± .027 & .976 ± .010 & .894 ± .007 & .977 ± .002 \\
 N-Flash & .869 ± .022 & .969 ± .008 & .882 ± .022 & .972 ± .008 \\
 FULL & \textbf{.900 ± .011} & \textbf{.979 ± .004} & \textbf{.907 ± .013} & \textbf{.980 ± .004} \\
 \bottomrule
 \end{tabular}%
 \label{tab:ablation}%
\end{table}%

\begin{figure}[tb]
 \centering
 \begin{subfigure}[b]{0.15\textwidth}

 \includegraphics[width=\textwidth]{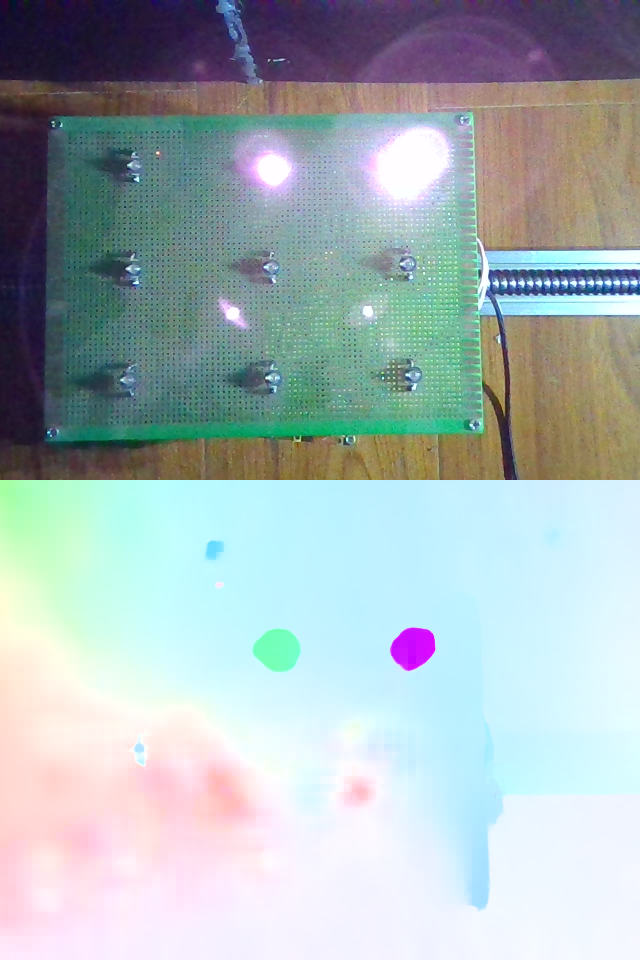}
 \caption{Without $\mathfrak{L}_{Vis}$}
 
 \end{subfigure}
 \hfill
 \begin{subfigure}[b]{0.15\textwidth}

 \includegraphics[width=\textwidth]{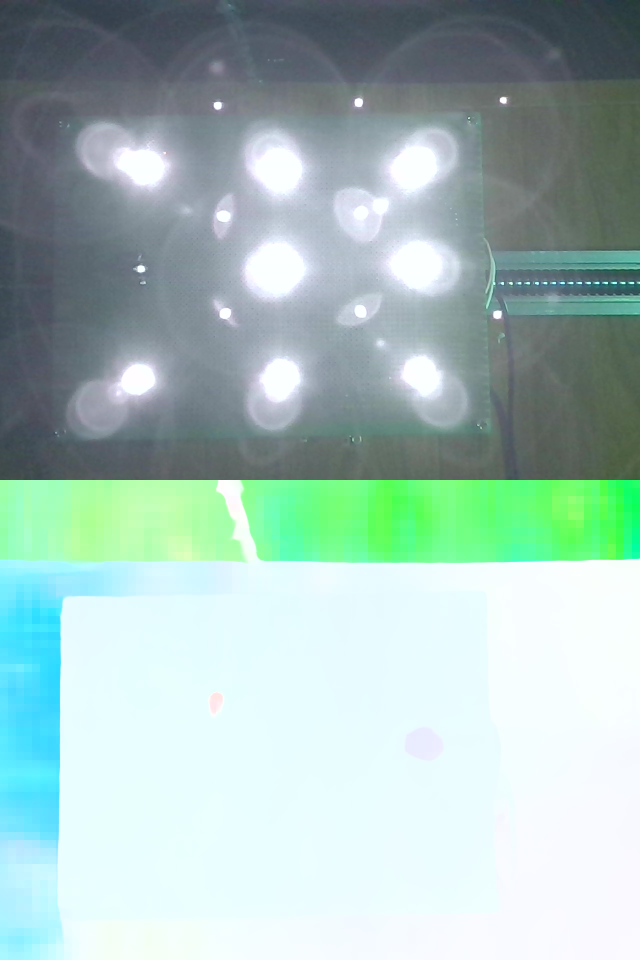}
 \caption{Without $\mathfrak{L}_{Per}$}
 \end{subfigure}
 \hfill
 \begin{subfigure}[b]{0.15\textwidth}

 \includegraphics[width=\textwidth]{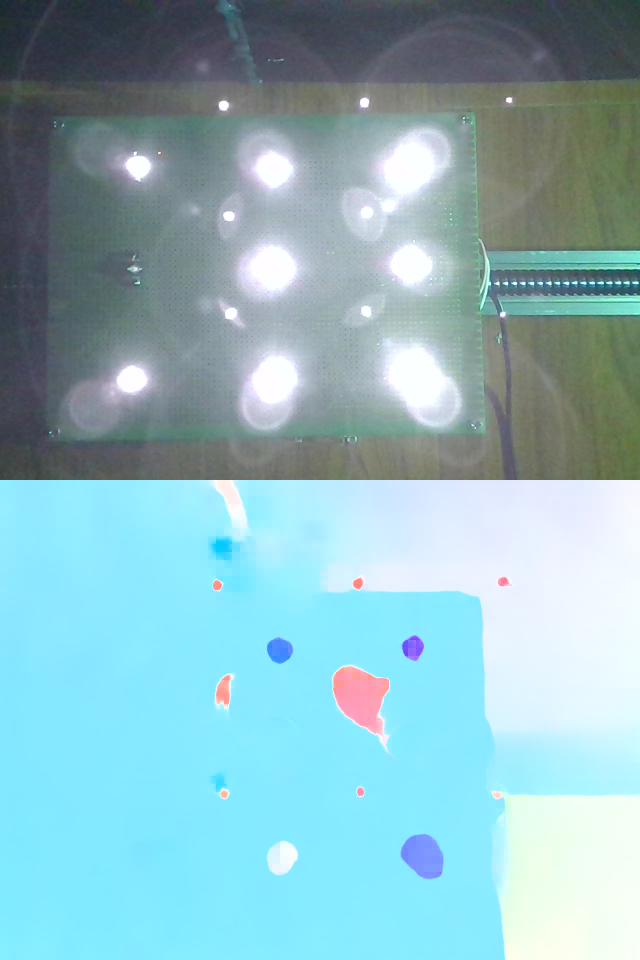}
 \caption{Without $\mathfrak{L}_{Flash}$}
 \end{subfigure}

 \caption{Disable one loss item when attacking RAFT.}
 \label{fig:validation}
\end{figure}

\begin{figure*}[tb]
 \centering
\includegraphics[width=1\textwidth]{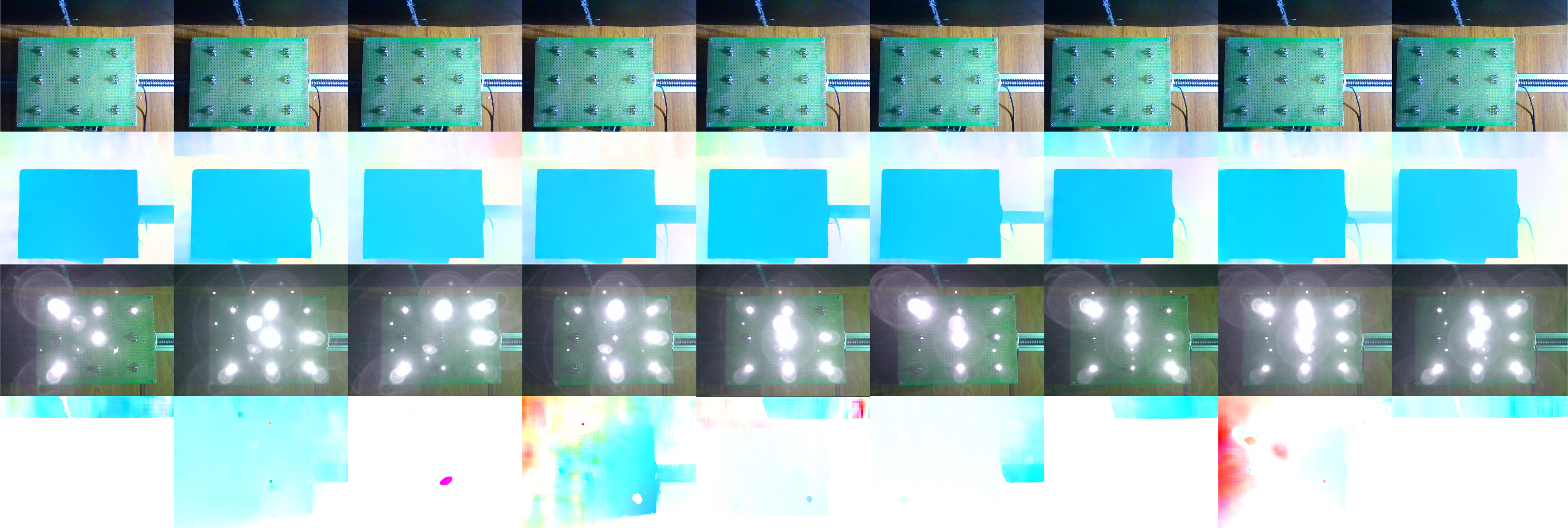}
 \caption{The target object is moving to the left from right. The original optical flow (up) and the attack results with full loss item (down) when attacking RAFT. }
 \label{fig:full_version}
\end{figure*}

From Table \ref{tab:ablation}, Fig. \ref{fig:validation} \ref{fig:full_version} shows the results, we observe the following effects when individual loss functions are removed: 
without invisibility loss, we could notice the AIV and AIP decreased by 1.16\%,	0.26\%. This indicates that the output optical flow has difficulty in accurately capturing the true motion. As shown in Fig. \ref{fig:validation}(a), the optical flow appears chaotic and disorganized.
Without imperceptibility loss, The AIV and AIV decreased by 1.70\%,	0.38\%. The optical flow tends to become blank, as seen in Fig. \ref{fig:validation}(b). This makes it challenging to locate and track the target object, as the flow lacks the necessary details to represent the object's motion accurately.
Without flash loss, both the AIV and AIP decreased by 3.22\%, 0.53\%. This reduction occurs because the absence of the flash loss diminishes the ability of the infrared lights to alter the lighting on the object's surface across multiple frames, so the general outline of the object remains visible, as shown in Fig. \ref{fig:validation}(c).
It can be observed that the full version of the method, which includes all three loss items, performs better than any of the incomplete versions. This indicates that each loss function contributes uniquely to achieving a stronger and more effective attack, and the flash loss is the most important item.

\subsection{Attack performance}
In this section, we test RILA under different physical conditions compared with Blank methods.

\subsubsection{Ambient Lighting}
OFENs rely on the light reflected from the surface of the target object, and our method attacks on this principle. Therefore, ambient lighting conditions are crucial, as it can interfere with the infrared light. This is a common scenario due to the natural variation in ambient lighting, such as the dimmer conditions in the evening and the brighter conditions at midday. To control these variables, we conducted our tests in a black box environment with four adjustable LED lights to simulate different ambient light conditions.

\begin{figure}[ht]
\centering

 \begin{subfigure}[b]{0.235\textwidth}

 \includegraphics[width=\textwidth]{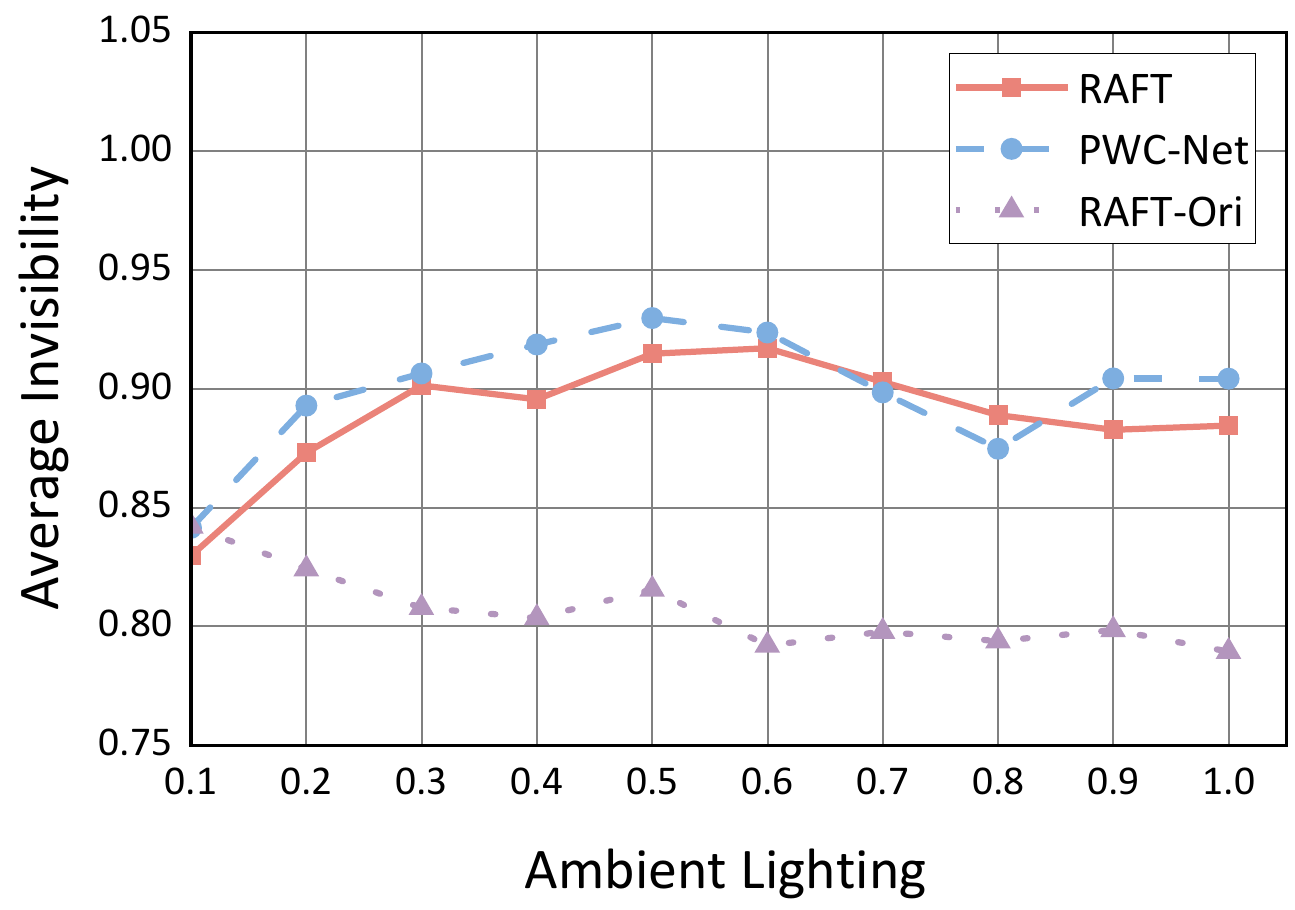}
 \caption{Without $\mathfrak{L}_{Vis}$}
 
 \end{subfigure}
  \begin{subfigure}[b]{0.235\textwidth}

 \includegraphics[width=\textwidth]{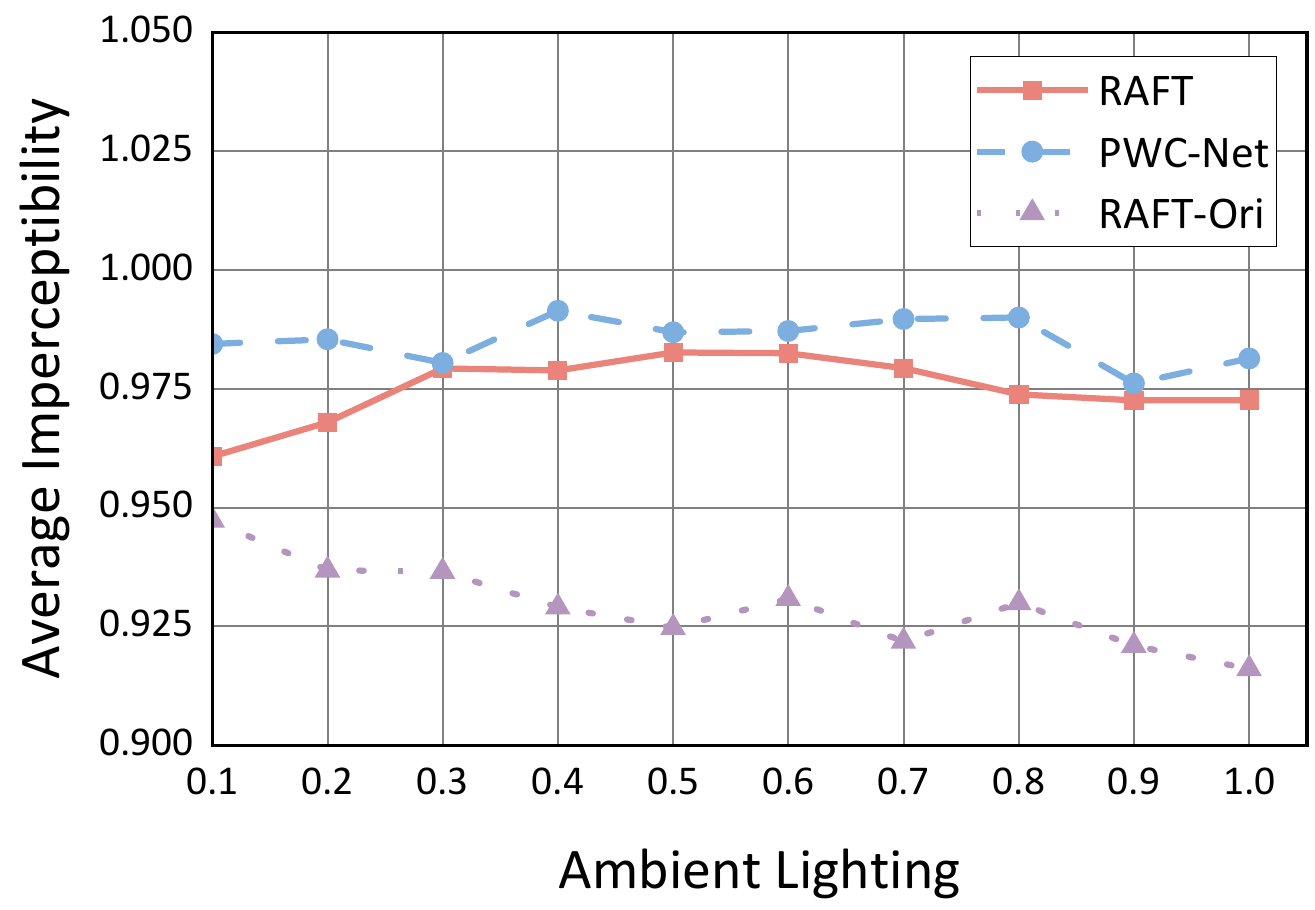}
 \caption{Without $\mathfrak{L}_{Vis}$}
 
 \end{subfigure}

\caption{The invisibility of RILA in different ambient lights.} 
\label{fig:lights_1}
\end{figure}


Figs. \ref{fig:lights_1}  show the AIV and AIP of RILA under different ambient lighting conditions. As the ambient light intensity increases, the original optical flow becomes clearer. When the ambient light intensity is below 0.3, the original optical flow is poorly estimated because OFENs struggle to track the light on the target object. As a result, the attacked flow becomes random, leading to a decline in invisibility and imperceptibility. In the range of 0.3 to 0.7 ambient light intensity, OFENs can accurately track the motion, and RILA effectively manipulates the OFENs to output a blank optical flow rather than a random flow, thus increasing the invisibility and imperceptibility. However, when the ambient light intensity exceeds 0.7, the enhanced natural light reduces the relative contribution of the infrared light on the object's surface, making the attack more challenging. Although the ambient lighting will affect our approach, the results indicate that RILA performs well across a wide range of lighting conditions.

\begin{table}[htbp]
 \centering
 \caption{Performance at different object motion speeds.}
 \begin{tabular}{c@{\hspace{2pt}}c@{\hspace{7pt}}c@{\hspace{7pt}}c@{\hspace{7pt}}c@{\hspace{0pt}}}
 \toprule
 \multirow{2}[4]{*}{\makecell{Speed\\ (cm/s)}} & \multicolumn{2}{c}{RAFT\cite{teed2020raft}} & \multicolumn{2}{c}{PWC-Net\cite{sun2018pwc}} \\
\cmidrule{2-5} & AIV & AIP & AIV & AIP \\
 \midrule
 6 & .929 ± .019 & .989 ± .005 & .932 ± .019 & .990 ± .005 \\
 9 & .908 ± .007 & .982 ± .005 & .911 ± .002 & .985 ± .008 \\
 12 & .899 ± .014 & .980 ± .005 & .888 ± .002 & .980 ± .006 \\
 15 & .868 ± .018 & .963 ± .005 & .879 ± .009 & .965 ± .001 \\
 18 & .859 ± .021 & .963 ± .013 & .873 ± .007 & .967 ± .008 \\
 21 & .866 ± .017 & .964 ± .010 & .863 ± .017 & .964 ± .010 \\
 24 & .859 ± .016 & .960 ± .012 & .872 ± .005 & .968 ± .003 \\
 27 & .856 ± .028 & .960 ± .016 & .875 ± .008 & .953 ± .012 \\
 30 & .858 ± .006 & .962 ± .005 & .859 ± .009 & .963 ± .007 \\
 33 & .838 ± .002 & .955 ± .010 & .842 ± .005 & .957 ± .010 \\
\bottomrule \end{tabular}%
 \label{tab:speed}%
\end{table}%

\subsubsection{Object Speed}
Considering the target object could have different speed in real world, and OFENs are based on the change of position of each pixel, It is more likely to track the optical flow when the change is huge and ignore the small motion, so we test our methods in different object speed.

\begin{table}[htbp]
 \centering
 \caption{Performance on different camera distances}
 \begin{tabular}{c@{\hspace{8pt}}c@{\hspace{8pt}}c@{\hspace{8pt}}c@{\hspace{8pt}}c@{\hspace{0pt}}}
 \toprule
 \multirow{2}[4]{*}{\makecell{Dis\\(cm)}} & \multicolumn{2}{c}{RAFT\cite{teed2020raft}} & \multicolumn{2}{c}{PWC-Net\cite{sun2018pwc}} \\
\cmidrule{2-5} & AIV & AIP & AIV & AIP \\
 \midrule
  30 & .896 ± .013 & .978 ± .005 & .900 ± .013 & .982 ± .005 \\
  40 & .892 ± .025 & .977 ± .008 & .909 ± .004 & .987 ± .008 \\
  50 & .867 ± .009 & .970 ± .004 & .871 ± .009 & .975 ± .003 \\
  60 & .858 ± .005 & .967 ± .002 & .860 ± .006 & .969 ± .004 \\
  70 & .857 ± .011 & .970 ± .003 & .866 ± .003 & .971 ± .002 \\
  80 & .880 ± .034 & .976 ± .012 & .899 ± .013 & .982 ± .005 \\
  90 & .859 ± .025 & .967 ± .010 & .858 ± .025 & .971 ± .010 \\
 100 & .881 ± .007 & .975 ± .002 & .885 ± .004 & .976 ± .001 \\
 110 & .872 ± .010 & .973 ± .003 & .874 ± .011 & .972 ± .001 \\
 120 & .853 ± .024 & .964 ± .005 & .868 ± .012 & .968 ± .005 \\
 \bottomrule
 \end{tabular}%
 \label{tab:distance}%
\end{table}%

Table \ref{tab:speed} shows the performance of our methods when attacking target objects moving at different speeds. We can observe that the AIV and AIP gradually decrease as the speed of the target object increases. This trend indicates that our method performs more effectively under lower object speeds. 

\subsubsection{Camera Distance}
Given that the distance between the target object and the camera varies in typical scenarios, it can affect the size of the target object in the input frame. Since OFENs are generally more adept at tracking large objects rather than small ones, we tested RILA at different camera distances. Due to the limitations of our testing environment, we selected eleven close distances, and the results are shown in Table \ref{tab:distance}.

We observe that the AIV and AIP do not change a lot with the increase of camera distance when attacking both models. This may be because the size of the area affected by the infrared light is fixed. Even if the object is far or near the camera, the infrared lights still fully cover the object's surface; this feature makes our methods could be used in wide environments.

\textcolor{revised}{
\subsubsection{Iteration number in RAFT}
Unlike traditional OFENs, RAFT employs an iterative Long Short-Term Memory (LSTM) module for decoding the optical flow, which typically results in improved performance as the Number Of Iterations (NOI) increases. To evaluate the robustness of our attack method, we conducted experiments across various iteration numbers. The results are summarized in Table \ref{tab:iteration_number}.
\begin{table}[htbp]
 \centering
 \caption{The performance in differ iteration number of RAFT.}
 \begin{tabular}{c@{\hspace{8pt}}c@{\hspace{8pt}}c@{\hspace{8pt}}c@{\hspace{8pt}}c@{\hspace{0pt}}}
 \toprule
 \multirow{2}[4]{*}{ \makecell{NOI}} & \multicolumn{2}{c}{Under Attacking} & \multicolumn{2}{c}{Original} \\
\cmidrule{2-5} & AIV & AIP & AIV & AIP \\
 \midrule
    5     & .935 ± .004 & .990 ± .009 & .900 ± .019 & .971 ± .012 \\
    6     & .925 ± .001 & .987 ± .003 & .854 ± .013 & .953 ± .016 \\
    7     & .933 ± .016 & .990 ± .003 & .841 ± .016 & .941 ± .003 \\
    8     & .941 ± .014 & .990 ± .005 & .839 ± .012 & .938 ± .018 \\
    9     & .935 ± .017 & .988 ± .005 & .813 ± .006 & .919 ± .007 \\
    10    & .919 ± .015 & .984 ± .004 & .828 ± .011 & .930 ± .003 \\
    11    & .923 ± .016 & .986 ± .015 & .819 ± .015 & .922 ± .011 \\
    12    & .919 ± .014 & .982 ± .004 & .820 ± .007 & .924 ± .018 \\
    13    & .916 ± .003 & .980 ± .018 & .815 ± .016 & .921 ± .018 \\
    14    & .915 ± .002 & .987 ± .014 & .819 ± .002 & .921 ± .012 \\
    15    & .912 ± .014 & .989 ± .005 & .814 ± .015 & .918 ± .018 \\
 \bottomrule
 \end{tabular}%
 \label{tab:iteration_number}%
\end{table}%
In the context of the original optical flow, we observed a significant decline in both AIV and AIP when the iteration count increased from 5 to 9. However, beyond 9 iterations, the AIV and AIP metrics remained relatively stable, indicating that additional iterations did not substantially improve performance.
When the victim model was subjected to our attack, a similar trend was noted: a notable drop in AIV and AIP occurred between 5 and 9 iterations, after which the metrics stabilized. Specifically, under attack, the AIV and AIP metrics decreased by only 2.48\% and 0.94\%, respectively, compared to a 9.52\% and 5.84\% reduction in the original optical flow. This suggests that our method maintains its effectiveness across different iteration numbers, demonstrating its robustness against the varying computational complexity of the RAFT model.
}
\subsection{Discussion}
From the above experiments, the proposed method has achieved high attack performance across different situations. It can be attributed to designing several bespoke loss functions to guide the generation of adversarial perturbations: Invisibility loss $\mathfrak{L}_{Vis}$ reduces the probability that the target object is detected. By maximizing it, the optical flow becomes blank, making it difficult for the system to detect any moving objects. Imperceptibility loss $\mathfrak{L}_{Per}$ ensures that the downstream tasks cannot correctly recognize the output of OFENs. A high $\mathfrak{L}_{Per}$ value indicates that the optical flow is unable to accurately perceive the motion, thereby degrading the performance of subsequent tasks. Flash loss $\mathfrak{L}_{Flash}$ leverages prior knowledge to maximize the difference between pixel points in consecutive frames. By maximizing $\mathfrak{L}_{Flash}$, the generated AEs create a significant gap between the pixel values of the current and previous frames, disrupting the optical flow estimation further. 

\section{Conclusion and Future Work}
\textcolor{revised}{
In this work, we have developed a practical real-time infrared light attack method in the physical world. 
Firstly, to ensure that the AEs are not noticeable by humans, we used infrared lights as the attack tool. Humans are not sensitive to infrared light, making this method highly stealthy and imperceptible. 
Considering physical attack is quite different from the  digital world, instead of adopting simulation experiments, we directly attack the target model using GA which do not require gradient information in the physical environment, ensuring that the generated AEs are robust to unexpected noise. 
To address the time-consuming nature of query-based methods, we designed a generative network that can produce AEs in real-time. 
Finally, we established a physical test environment consisting of tracks and controllers to simulate the movement of target object, ensuring that the AEs are tested and validated in a realistic setting.
Experiments demonstrate that our methods successfully attack the target system in complex environments, including varying object speeds, different ambient lighting conditions, and varying camera distances.
This research shows the infrared lights attack has a great adaptive capacity to disturb the OFEN at different environment, which shows a heavy threaten to the current OFEN-driven security system. 
}

For future works, considering that autonomous driving systems rely not only on optical flow estimation but also on other tasks such as object detection, we plan to extend our work to attack models in these additional tasks. This will enable us to achieve a more comprehensive and robust multimodal attack, further enhancing the effectiveness of our approach. By addressing these areas, we aim to improve the efficiency of our adversarial attack methods, making them more applicable to a wide range of real-world scenarios.

{
    \small
    \bibliographystyle{ieeenat_fullname}
    \bibliography{main}

@String(CVPR= {IEEE Conf. Comput. Vis. Pattern Recog.})

@String(AAAI = {AAAI})

@String(CVPR  = {CVPR})

@inproceedings{ranjan2019attacking,
  title={Attacking optical flow},
  author={Ranjan, Anurag and Janai, Joel and Geiger, Andreas and Black, Michael J},
  booktitle={Proceedings of the IEEE/CVF international conference on computer vision},
  pages={2404--2413},
  year={2019}
}

@inproceedings{gnanasambandam2021optical,
  title={Optical adversarial attack},
  author={Gnanasambandam, Abhiram and Sherman, Alex M and Chan, Stanley H},
  booktitle={Proceedings of the IEEE/CVF International Conference on Computer Vision},
  pages={92--101},
  year={2021}
}

@inproceedings{capito2020optical,
  title={Optical flow based visual potential field for autonomous driving},
  author={Capito, Linda and Ozguner, Umit and Redmill, Keith},
  booktitle={2020 IEEE Intelligent Vehicles Symposium (IV)},
  pages={885--891},
  year={2020},
  organization={IEEE}
}

@inproceedings{dosovitskiy2015flownet,
  title={Flownet: Learning optical flow with convolutional networks},
  author={Dosovitskiy, Alexey and Fischer, Philipp and Ilg, Eddy and Hausser, Philip and Hazirbas, Caner and Golkov, Vladimir and Van Der Smagt, Patrick and Cremers, Daniel and Brox, Thomas},
  booktitle={Proceedings of the IEEE international conference on computer vision},
  pages={2758--2766},
  year={2015}
}

@inproceedings{sun2018pwc,
  title={Pwc-net: Cnns for optical flow using pyramid, warping, and cost volume},
  author={Sun, Deqing and Yang, Xiaodong and Liu, Ming-Yu and Kautz, Jan},
  booktitle={Proceedings of the IEEE conference on computer vision and pattern recognition},
  pages={8934--8943},
  year={2018}
}

@inproceedings{teed2020raft,
  title={Raft: Recurrent all-pairs field transforms for optical flow},
  author={Teed, Zachary and Deng, Jia},
  booktitle={Computer Vision--ECCV 2020: 16th European Conference, Glasgow, UK, August 23--28, 2020, Proceedings, Part II 16},
  pages={402--419},
  year={2020},
  organization={Springer}
}

@inproceedings{narodytska2017simple,
  title={Simple Black-Box Adversarial Attacks on Deep Neural Networks.},
  author={Narodytska, Nina and Kasiviswanathan, Shiva Prasad},
  booktitle={CVPR Workshops},
  volume={2},
  number={2},
  year={2017}
}

@inproceedings{croce2019sparse,
  title={Sparse and imperceivable adversarial attacks},
  author={Croce, Francesco and Hein, Matthias},
  booktitle={Proceedings of the IEEE/CVF international conference on computer vision},
  pages={4724--4732},
  year={2019}
}

@article{chamorro2006new,
  title={A new approach to motion pattern recognition and its application to optical flow estimation},
  author={Chamorro-Martinez, Jes{\'u}s and Fern{\'a}ndez-Valdivia, Joaqu{\'\i}n},
  journal={IEEE Transactions on Systems, Man, and Cybernetics, Part C (Applications and Reviews)},
  volume={37},
  number={1},
  pages={39--51},
  year={2006},
  publisher={IEEE}
}

@article{sivaraman2013looking,
  title={Looking at vehicles on the road: A survey of vision-based vehicle detection, tracking, and behavior analysis},
  author={Sivaraman, Sayanan and Trivedi, Mohan Manubhai},
  journal={IEEE transactions on intelligent transportation systems},
  volume={14},
  number={4},
  pages={1773--1795},
  year={2013},
  publisher={IEEE}
}

@article{haag1999combination,
  title={Combination of edge element and optical flow estimates for 3D-model-based vehicle tracking in traffic image sequences},
  author={Haag, Michael and Nagel, Hans-Hellmut},
  journal={International Journal of Computer Vision},
  volume={35},
  pages={295--319},
  year={1999},
  publisher={Springer}
}

@inproceedings{song2001fast,
  title={Fast optical flow estimation and its application to real-time obstacle avoidance},
  author={Song, Kai-Tai and Huang, Jui-Hsiang},
  booktitle={Proceedings 2001 ICRA. IEEE International Conference on Robotics and Automation (Cat. No. 01CH37164)},
  volume={3},
  pages={2891--2896},
  year={2001},
  organization={IEEE}
}

@inproceedings{wang2021can,
  title={I can see the light: Attacks on autonomous vehicles using invisible lights},
  author={Wang, Wei and Yao, Yao and Liu, Xin and Li, Xiang and Hao, Pei and Zhu, Ting},
  booktitle={Proceedings of the 2021 ACM SIGSAC Conference on Computer and Communications Security},
  pages={1930--1944},
  year={2021}
}

@inproceedings{eykholt2018robust,
  title={Robust physical-world attacks on deep learning visual classification},
  author={Eykholt, Kevin and Evtimov, Ivan and Fernandes, Earlence and Li, Bo and Rahmati, Amir and Xiao, Chaowei and Prakash, Atul and Kohno, Tadayoshi and Song, Dawn},
  booktitle={Proceedings of the IEEE conference on computer vision and pattern recognition},
  pages={1625--1634},
  year={2018}
}

@article{zheng2023robust,
  title={Robust physical-world attacks on face recognition},
  author={Zheng, Xin and Fan, Yanbo and Wu, Baoyuan and Zhang, Yong and Wang, Jue and Pan, Shirui},
  journal={Pattern Recognition},
  volume={133},
  pages={109009},
  year={2023},
  publisher={Elsevier}
}

@inproceedings{hu2021naturalistic,
  title={Naturalistic physical adversarial patch for object detectors},
  author={Hu, Yu-Chih-Tuan and Kung, et al.},
  booktitle={Proceedings of the IEEE/CVF International Conference on Computer Vision},
  pages={7848--7857},
  year={2021}
}

@article{lee2019physical,
  title={On physical adversarial patches for object detection},
  author={Lee, Mark and Kolter, Zico},
  journal={arXiv preprint arXiv:1906.11897},
  year={2019}
}

@article{brown2017adversarial,
  title={Adversarial patch},
  author={Brown, Tom B and Man{\'e}, Dandelion and Roy, Aurko and Abadi, Mart{\'\i}n and Gilmer, Justin},
  journal={arXiv preprint arXiv:1712.09665},
  year={2017}
}

@inproceedings{liu2019perceptual,
  title={Perceptual-sensitive gan for generating adversarial patches},
  author={Liu, Aishan and Liu, Xianglong and Fan, Jiaxin and Ma, Yuqing and Zhang, Anlan and Xie, Huiyuan and Tao, Dacheng},
  booktitle={Proceedings of the AAAI conference on artificial intelligence},
  volume={33},
  number={01},
  pages={1028--1035},
  year={2019}
}

@inproceedings{wang2023rfla,
  title={Rfla: A stealthy reflected light adversarial attack in the physical world},
  author={Wang, Donghua and Yao, Wen and Jiang, Tingsong and Li, Chao and Chen, Xiaoqian},
  booktitle={Proceedings of the IEEE/CVF international conference on computer vision},
  pages={4455--4465},
  year={2023}
}

@inproceedings{duan2021adversarial,
  title={Adversarial laser beam: Effective physical-world attack to dnns in a blink},
  author={Duan, Ranjie and Mao, Xiaofeng and Qin, A Kai and Chen, Yuefeng and Ye, Shaokai and He, Yuan and Yang, Yun},
  booktitle={Proceedings of the IEEE/CVF conference on computer vision and pattern recognition},
  pages={16062--16071},
  year={2021}
}

@inproceedings{guo2024invisible,
  title={Invisible optical adversarial stripes on traffic sign against autonomous vehicles},
  author={Guo, Dongfang and Wu, Yuting and Dai, Yimin and Zhou, Pengfei and Lou, Xin and Tan, Rui},
  booktitle={Proceedings of the 22nd Annual International Conference on Mobile Systems, Applications and Services},
  pages={534--546},
  year={2024}
}

@inproceedings{bhupathiraju2024vulnerability,
  title={On the Vulnerability of Traffic Light Recognition Systems to Laser Illumination Attacks},
  author={Bhupathiraju, Sri Hrushikesh Varma and Sugawara, Takeshi and Sato, Takami and Chen, Qi Alfred and Clifford, Michael and Rampazzi, Sara},
  booktitle={ISOC Symposium on Vehicle Security and Privacy (VehicleSec). ISOC, San Diego, CA, USA. https://doi. org/10},
  volume={14722},
  year={2024}
}
}


\end{document}